\documentclass[10pt,twocolumn,letterpaper]{article}

\usepackage{iccv}
\usepackage{times}
\usepackage{epsfig}
\usepackage{graphicx}
\usepackage{amsmath}
\usepackage{amssymb}
\usepackage{booktabs}
\usepackage{xcolor}
\usepackage[accsupp]{axessibility}

\usepackage[breaklinks=true,bookmarks=false]{hyperref}

\iccvfinalcopy % *** Uncomment this line for the final submission

\def\ie{\emph{i.e}\onedot}

\def\rvs{{\mathbf{s}}}
\def\rvt{{\mathbf{t}}}

\def\rvw{{\mathbf{w}}}

\def\rmR{{\mathbf{R}}}

\ificcvfinal\fi

\begin{document}

%%%%%%%%% TITLE
\title{GVP: Generative Volumetric Primitives}

\author{Mallikarjun B R~~~Xingang Pan~~~Mohamed Elgharib~~~Christian Theobalt\\ 
	Max Planck Insitute for Informatics,~~~Saarland Informatics Campus
}
\maketitle
% Remove page # from the first page of camera-ready.
\ificcvfinal\thispagestyle{empty}\fi

%%%%%%%%% ABSTRACT
\begin{abstract}

   Advances in 3D-aware generative models have pushed the boundary of image synthesis with explicit camera control.
   To achieve high-resolution image synthesis, several attempts have been made to design efficient generators, such as hybrid architectures with both 3D and 2D components.
   However, such a design compromises multiview consistency, and the design of a pure 3D generator with high resolution is still an open problem.
   In this work, we present Generative Volumetric Primitives (GVP), the first pure 3D generative model that can sample and render 512-resolution images in real-time.
   GVP jointly models a number of volumetric primitives and their spatial information, both of which can be efficiently generated via a 2D convolutional network.
   The mixture of these primitives naturally captures the sparsity and correspondence in the 3D volume.
   The training of such a generator with high degree of freedom is made possible through a knowledge distillation technique.
   Experiments on several datasets demonstrate superior efficiency and 3D consistency of GVP over the state-of-the-art.
   
\end{abstract}

%%%%%%%%% BODY TEXT
\section{Introduction}

Synthesizing photorealistic objects with high-resolution and multi-view consistency is a long-term goal in computer vision and graphics.
Recently, advances in 3D-aware generative adversarial networks (GANs) have made inspiring progress towards this goal~\cite{Schwarz2020NEURIPS,piGAN2021,Niemeyer2020GIRAFFE,gu2022stylenerf,Chan2022}.
By learning from unstructured 2D images, 3D-aware GANs can synthesize different views of generated objects with explicit control over camera pose.
This is made possible through 3D generator architectures that naturally possess a 3D inductive bias.

The key challenge towards a high-quality 3D-aware GAN lies in the design of an efficient 3D generator.
For example, voxel-based 3D representation~\cite{henzler2019escaping} suffers from cubic memory growth,
 and thus can only afford a limited resolution.
Recently, a more widely used representation for 3D-aware GAN is neural radiance fields (NeRF)~\cite{mildenhall2020nerf,Schwarz2020NEURIPS,piGAN2021}, which integrates a coordinate-based MLP and volume rendering.
As NeRF requires querying the MLP densely across 3D space, it is prohibitively computationally expensive for high-resolution image rendering.
Several works propose to improve efficiency via a compromised hybrid generator that models a 3D feature space via NeRF followed by a 2D CNN-based renderer that upsamples the feature to a high-resolution 2D image~\cite{Niemeyer2020GIRAFFE,gu2022stylenerf,Chan2022,orel2022styleSDF}.
However, the use of a 2D CNN compromises 3D consistency and is vulnerable to artifacts for camera views unseen during training.
Although a few recent works attempt to improve the efficiency of a pure 3D generative model~\cite{epigraf,Schwarz2022NEURIPS}, the synthesis of high-resolution images in real-time remains an open problem. % (\eg, 512)

\begin{figure}
\centering
\includegraphics[width=0.5\textwidth]{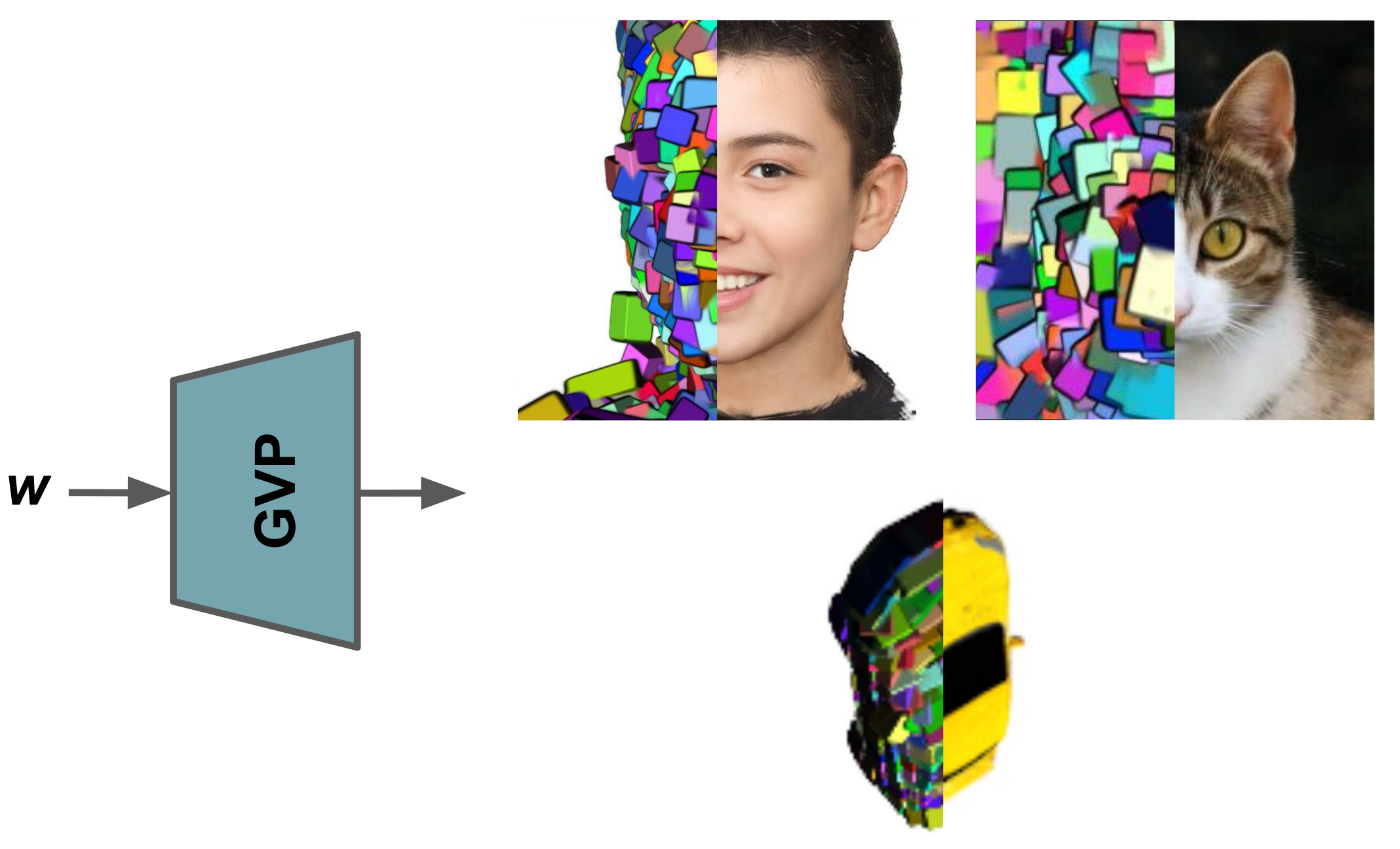}
%\vspace{-0.5cm}
\caption{We propose the first 3D-aware generative model that relies on pure 3D representation and can render images at a resolution of $512 \times 512$ in real-time. At the core of our method, a scene is represented by a number of volumetric primitives that can be rendered very efficiently. Our method can generalize to different object categories such as human and cat portraits and cars.}
%\vspace{-0.1cm}
\label{fig:teaser}
\end{figure}

In this work, we present Generative Volumetric Primitives (GVP), which, to the best of our knowledge, is the first 3D-aware generative model that is based on a pure 3D representation and can sample and render 512-resolution images in real-time.
To achieve high rendering efficiency, we take the first attempt to build a 3D generative model based on the mixture of volumetric primitives (MVP) representation~\cite{lombardi2021mixture}.
Each primitive models the color and density of a local volume and a number of primitives are composed together to form the entire 3D volume, as shown in Fig.~\ref{fig:teaser}.
Unlike the original MVP that heavily relies on mesh tracking to determine the spatial information (\ie, position, orientation, and scale) of the primitives, our generator learns to model the volumetric primitives and their spatial information jointly.
Such a representation can be efficiently modeled via a 2D convolutional network architecture and naturally captures spatial sparsity, thus is extremely efficient to sample and render compared to previous 3D-aware GANs.

The use of multiple movable primitives also introduces a high degree of freedom to our generator, making it unstable to train via a conventional adversarial loss on the raw training images.
To address this issue, we propose to bypass the adversarial training via knowledge distillation of a pretrained 3D-aware GAN, EG3D~\cite{Chan2022}.
Such knowledge distillation significantly stabilizes training and inherits the generative capability of EG3D.
Although EG3D has inconsistent rendering for out-of-distribution camera views, we found the in-distribution views sufficient to train our model with high quality, which would then be able to perform consistent rendering for any camera views.

We conduct an extensive evaluation of GVP on several datasets including human face, cat face, and cars.
Thanks to the highly efficient design, GVP achieves much faster rendering than previous pure 3D GANs and preserves better multiview consistency compared with those with hybrid architectures.
Visualization of the learned primitives demonstrates that GVP effectively captures the sparsity of the 3D volume, avoiding redundant computation and memory costs.
The predicted positions of the volumetric primitives also adapt to different samples, which to some extent captures correspondence between samples.

%%%%%%%%% BODY TEXT
\section{Related works}

Building generative models is a long-standing computer vision problem. Early efforts include works such as learning 3D Morphable models of the human face~\cite{Blanz99,Paysan09}. These approaches, however, lack photorealism and are trained using controlled data captured with multi-view camera rigs. With the advances in deep learning, several generative models have been proposed~\cite{Karras19,Karras2019stylegan2}. These models are learned purely from in-the-wild 2D images and produce highly photorealistic results. However, they are 2D-based and hence produce results that are not multi-view consistent. The past couple of years witnessed a rapidly growing interest in 3D generative models. This is largely fueled with the recent advances in implicit scene representations and Neural Radiance Fields (NeRF) \cite{Park_2019_CVPR,mildenhall2020nerf} which quickly revolutionized 3D scene understanding and rendering in an unprecedented manner. While there are several 3D generative methods designed for a specific class of objects such as the human face~\cite{Cao22}, body~\cite{2021narf} and hands~\cite{corona2022lisa}, several other approaches are generic~\cite{piGAN2021,Schwarz2022NEURIPS,deng2022gram,Chan2022,bergman2022gnarf,Xiang22}. These approaches combine volumetric rendering with adversarial training to achieve highly photorealistic renderings that are multi-view consistent. 

The works of Chan~\etal~\cite{piGAN2021} and Schwarz~\etal~\cite{Schwarz2020NEURIPS} were one of the first to show how to marry NeRF representation with adversarial training. They use co-ordinate based MLP to parameterize their generator and train in an adversarial manner. As they need to densely sample space to render the scene, it is quite slow. VoxGraf~\cite{Schwarz2022NEURIPS} acknowledges the fact that volumetric rendering is usually very slow due to querying multiple points along rays. To address this limitation, VoxGraf utilizes sparse voxel grid representations. This improves computational efficiency, but still, their method is not real-time even at $256\times256$ resolution. Deng~\etal~\cite{deng2022gram} proposed an alternative solution to handle the large computational requirements of volumetric rendering. Their approach uses 2D manifolds to guide point sampling and radiance field learning. These manifolds are embedded in the 3D volume and help in reducing the computational complexity while still producing multi-view consistent results. However, such manifolds can lead to clear rendering artifacts, especially in side-view angles, where the manifolds are almost parallel to camera rays. 
Rebain~\etal~\cite{Rebain22} showed it is possible to learn a generative model using single-view in-the-wild 2D images without adversarial training. They follow an autodecoder architecture that learns a shared latent representation. The method is trained in a self-supervised manner, where an off-the-shelf 2D landmark detector is used to determine the camera poses. While most of these methods~\cite{piGAN2021,Schwarz2022NEURIPS} and more~\cite{xu2021volumegan,epigraf} produce photorealistic renderings that are 3D-consistent, they suffer from expensive computational requirements that prohibit them from rendering in real-time.  

\begin{figure*}
\centering
\includegraphics[width=1.0\textwidth]{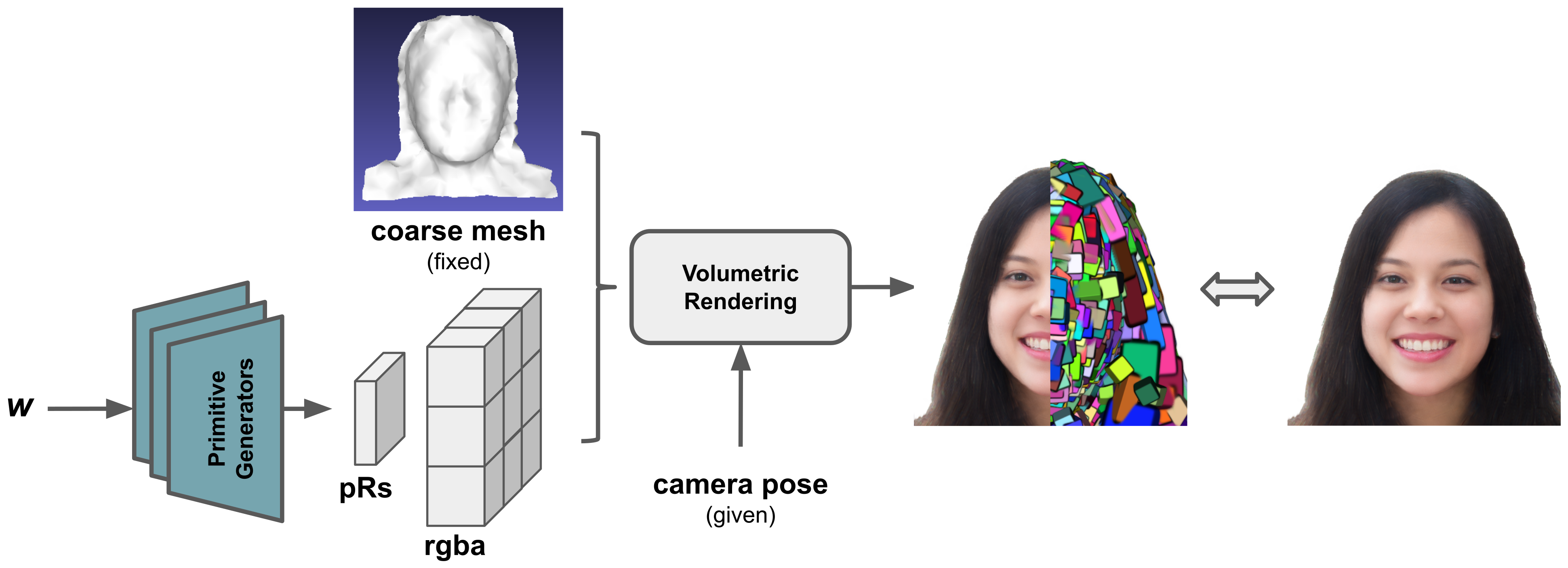}
%\vspace{-0.5cm}
\caption{\textbf{Method overview:} We propose a generative model that represents each sample using a number of volumetric primitives (voxel grid) that can be rendered very efficiently in real-time. Our method consists of 3 2D-CNN-based generators. The geometric generator takes latent vector $w$ and outputs the delta position, delta orientation, and delta scales of primitives with respect to initialized position, orientation, and scale using a coarse mesh. The output of the geometric generator is represented as pRs. The second generator takes the latent vector $w$ as input and outputs the density values of all the primitives at once. And the last generator takes latent vector $w$ and viewing direction as input and outputs color content of the primitives. The density and color outputs are jointly represented as rgba. Note we use pre-trained mapping network of EG3D~\cite{Chan2022} to map the gaussian noise to our latent vector $w$. The scene can then be volumetrically rendered efficiently to synthesize samples under any camera pose.}
%\vspace{-0.1cm}
\label{fig:pipeline}
\end{figure*}

To address the expensive computational requirement of the previously discussed generative models, another class of methods performs volumetric rendering in a low-resolution instead~\cite{Niemeyer2020GIRAFFE,Chan2022,gu2022stylenerf,orel2022styleSDF,bergman2022gnarf,Zhang22}. This is then followed by a super-resolution network that upscales the output to the full resolution. An example is the method EG3D by Chan~\etal~\cite{Chan2022}. This method extracts 2D StyleGAN2 features~\cite{Karras2019stylegan2} and utilizes a 3D tri-planar representation together with a lightweight feature decoder to extract 3D features. A volumetric renderer operating in a low resolution produces an image output, which is then passed through the super-resolution network. The method is trained in an adversarial manner using in-the-wild 2D images. 
EG3D~\cite{Chan2022} produces impressive multi-view consistent photorealistic results and is significantly more computationally efficient than previously discussed methods~\cite{piGAN2021,Schwarz2022NEURIPS,xu2021volumegan,epigraf}. This super-resolution-based framework has also been utilized in several other works~\cite{gu2022stylenerf,orel2022styleSDF,bergman2022gnarf,Zhang22}, including methods that handle articulated objects~\cite{bergman2022gnarf} and others that use an SDF representation~\cite{orel2022styleSDF}. However, the framework suffers from one main limitation, the fact that it is not real-3D. While the super-resolution network helps in reducing computational complexity significantly, it doesn't guarantee multi-view consistent results. Hence, the framework fails when rendering from viewpoints that are out of the training distribution.

\section{Method}

%%%%%%%%% BODY TEXT

The goal of our method is to build efficient generative volumetric models. 
We take inspiration from the recent method, MVP~\cite{lombardi2021mixture} to represent the 3D scene.
In specific, MVP~\cite{lombardi2021mixture} proposes an efficient 3D scene representation with a mixture of volumetric primitives. 
But to train MVP, they need registered meshes of an identity performing diverse expressions.
Our goal is to build and generalize the training across different categories, where sophisticated 3D morphable models don't exist.
Here we show, it's possible to build a generative model without the need for sophisticated morphable models for different categories.
One way to train this framework is by learning the whole model end-to-end in an adversarial manner using only loss from an image-based discriminator similar to existing methods~\cite{piGAN2021}.
Our initial experiments in that direction proved to be unstable because of multiple moving components.
The overview of our method can be found in Fig.~\ref{fig:pipeline}
To address this issue, we trained our model in a knowledge-distillation fashion using pre-trained EG3D~\cite{Chan2022} models.
In the following, we describe the representation of our model in detail in Sec.~\ref{subsec:representation} and also provide information about training in Sec.~\ref{subsec:training}.

\subsection{Representation}
\label{subsec:representation}
\paragraph{Volumetric Primitives:} 
As mentioned before, we take inspiration from MVP~\cite{lombardi2021mixture} to model our scene efficiently.
The efficiency arises because of mainly $2$ reasons. One, by having only sparsely distributed volumetric primitives in the scene, they can avoid sampling unnecessary regions that do not contribute to volumetric rendering. Second, the content of the primitives can be generated at once by an efficient CNN network instead of multiple queries required by a co-ordinate based MLP methods~\cite{piGAN2021}. 
As we want to learn a generative model for a category, let $\rvw \in \mathbb{R}^{512}$ be the latent vector that defines a sample.
Let $N_{prim}$ be the number of primitives in the scene and each of the primitives is a 3D voxel grid representing only a small region in the 3D space. 
The position, orientation, and scale of $k^{th}$ primitive in 3D space are defined by $\rvt_k(\rvw) \in \mathbb{R}^3$, $\rmR_k(\rvw) \in$ SO(3), and $\rvs_k(\rvw) \in \mathbb{R}^3$ respectively. 
The content of each primitive is defined by a dense voxel grid $V_k(\rvw) \in \mathbb{R}^{4 \times M_x \times M_y \times M_z}$ that contains color and opacity information. In all our experiments, we use $M_x=M_y=M_z=32$.
The payload of the volumetric primitive contains color and opacity and is defined as a function of sample $\rvw$.

\begin{figure*}

\newcommand{\lspace}[6]{\hspace{0.3cm} #1 \hspace{1.4cm}  #2  \hspace{1.4cm} #3 \hspace{1.4cm} #4 \hspace{1.4cm} #5 \hspace{1.4cm} #6 }

\centering
\includegraphics[width=1.0\textwidth]{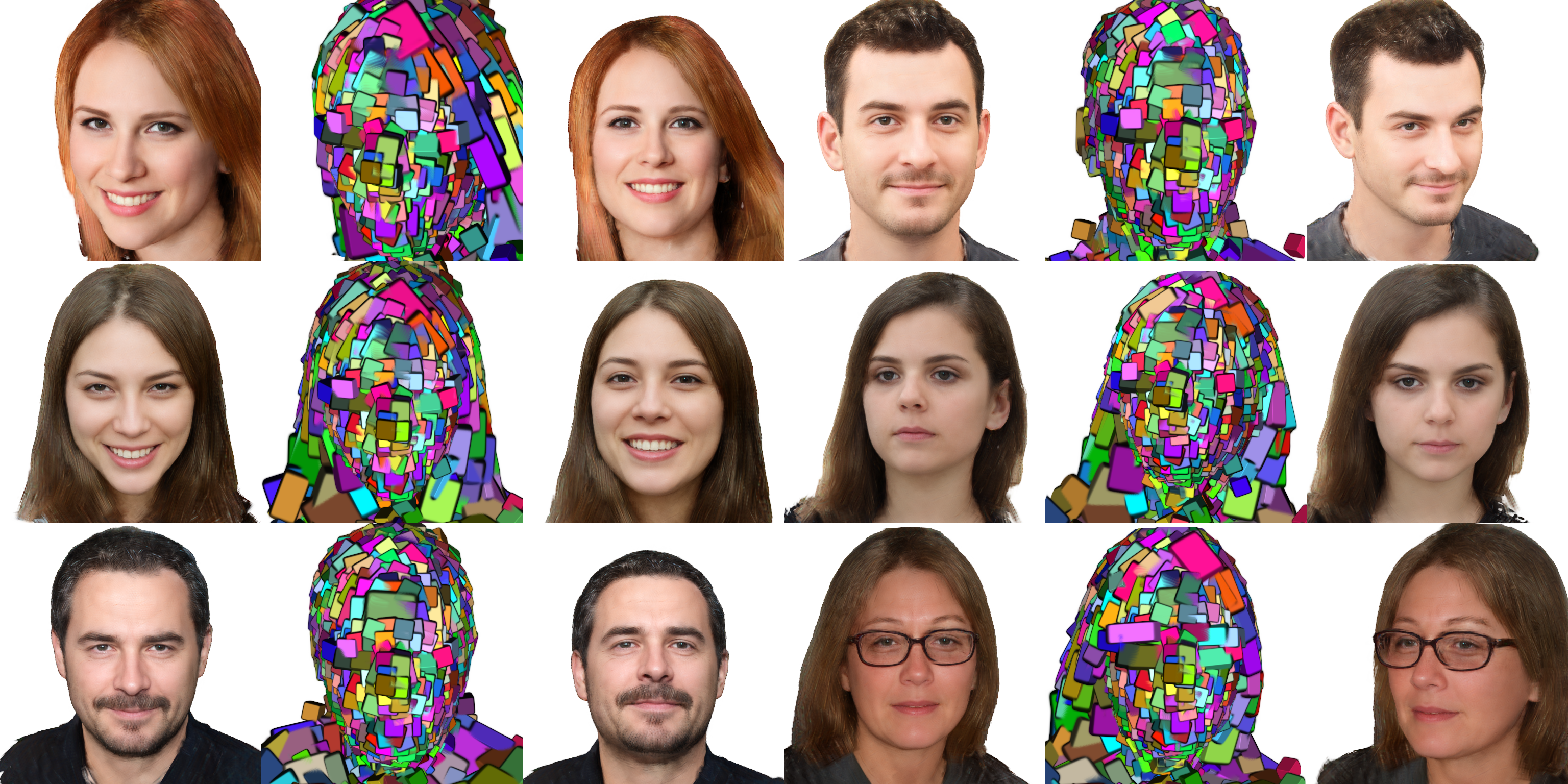}
%\vspace{-0.5cm}
\lspace{Sample}{Primitives}{Novel View}{Sample}{Primitives}{Novel View}

\caption{Here we show samples of our method trained on the FFHQ dataset. We also provide renderings of primitives. The color coding of primitives is consistent across different samples. Please note the semantically similar parts are occupied by the same colored primitives across different samples, which provides correspondence information. }
%\vspace{-0.1cm}
\label{fig:ours_ffhq}
\end{figure*}

\paragraph{Guide Mesh:}
MVP~\cite{lombardi2021mixture} utilizes tracked mesh to guide the position and orientation of primitives. Although it is possible to obtain tracked mesh for faces, obtaining the same for other categories like cats and cars becomes challenging.
Our surprising finding is that, using our training strategy, we do not need registered meshes across different samples of the category to train the primitives.
We make use of a fixed mesh as the guide mesh, $\mathcal{M}$ for all samples, and learn the delta position, orientation, and scales for all the primitives.
In order to get the fixed guide mesh $\mathcal{M}$, we extract a mesh from a manually chosen EG3D~\cite{Chan2022} generated sample  and build its UV map using Blender~\cite{blender}. We observe that the performance of our model does not depend much on the type of sample chosen.
We obtain the guide mesh's contribution of primitive position and orientation similar to MVP and let it be $\hat \rvt_k(\mathcal{M})$ and $\hat \rmR_k(\mathcal{M})$. We refer readers to MVP~\cite{lombardi2021mixture} for the exact details.
And we also learn delta position, orientation, and scales as a function of latent vector $\rvw$ to generalize to different samples.
The final position, orientation, and scale are defined by:
$$ \rvt_k(\rvw) = \hat \rvt_k(\mathcal{M}) +  \delta_{\rvt_k}(\rvw)  $$ 
$$\rmR_k(\rvw) = \hat \rmR_k(\mathcal{M}) \cdot  \delta_{\rmR_k}(\rvw) $$
$$\rvs_k(\rvw) = \hat \rvs_k +  \delta_{\rvs_k}(\rvw)
$$
We set $\hat \rvs_k$ to a fixed value for all primitives which was found empirically. 

\paragraph{Network Architecture:} 
Our model mainly consists of 3 generators.
The 3 generators are geometric generator $\mathcal{G}_{geo}$ and two payload generators $\mathcal{G}_{a},\mathcal{G}_{rgb}$.
The geometric generator $\mathcal{G}_{geo}:\mathbb{R}^{512} \rightarrow \mathbb{R}^{9\times N_{prim}}$ is for obtaining the delta position, orientation, and scale of the primitives.
And the payload generators $\mathcal{G}_{a} : \mathbb{R}^{512} \rightarrow \mathbb{R}^{1 \times M^3 \times N_{prim}}$ and $\mathcal{G}_{rgb} : \mathbb{R}^{512+3} \rightarrow \mathbb{R}^{3 \times M^3 \times N_{prim}}$ are used to obtain the opacity and color values.
\paragraph{Efficient Differentiable Rendering}
Once we have the primitives and their attributes, we make use of efficient differentiable rendering to render our scene. 
For a given ray $r_\mathbf{p}(t) =\mathbf{o}_\mathbf{p} + t \mathbf{d}_\mathbf{p}$ with starting position $\mathbf{o}_\mathbf{p}$ and ray direction $\mathbf{d}_\mathbf{p}$,
$$
\mathcal{I}_p = 
\int_{t_\text{min}}^{t_\text{max}}{
\mathbf{V}_{\text{col}}(r_\mathbf{p}(t)) \cdot \frac{dT(t)}{dt}\cdot dt}
\enspace .
$$
$$
T(t)=
\min\Big(
\int_{t_\text{min}}^{t}{
\mathbf{V}_{\alpha}(r_\mathbf{p}(t))\cdot dt
, 1\Big)}
\enspace .
$$
Here, $\mathbf{V}_{\text{col}}$ and $\mathbf{V}_{\alpha}$ are the  color and opacity values at a given point on the primitive.

\begin{figure*}

\newcommand{\lspace}[6]{\hspace{0.3cm} #1 \hspace{1.4cm}  #2  \hspace{1.4cm} #3 \hspace{1.4cm} #4 \hspace{1.4cm} #5 \hspace{1.4cm} #6 }

\centering
\includegraphics[width=1.0\textwidth]{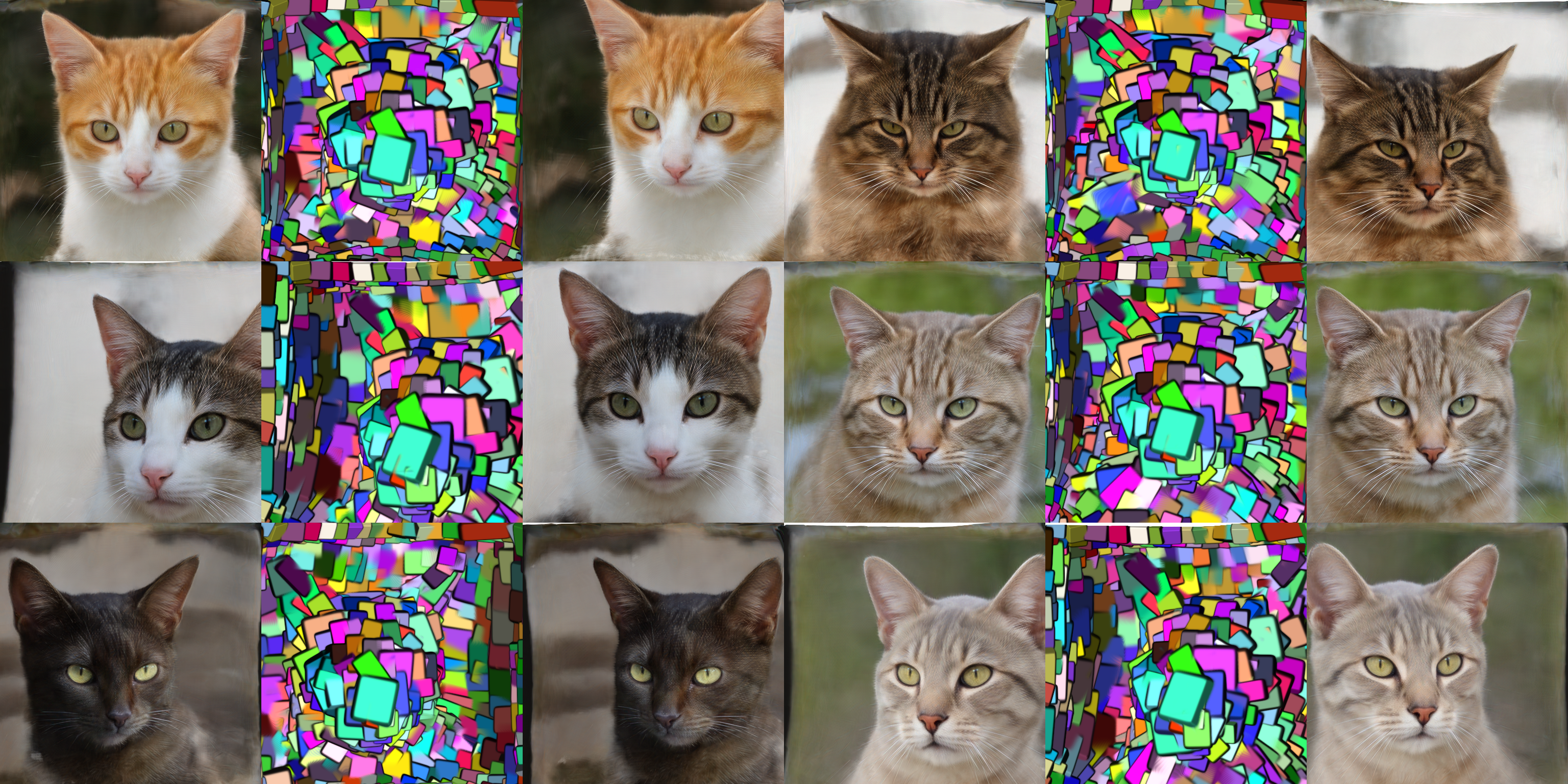}
%\vspace{-0.5cm}

\lspace{Sample}{Primitives}{Novel View}{Sample}{Primitives}{Novel View}

\vspace{0.1cm}
\caption{Here we show samples of our method trained on the AFHQ dataset. We also provide renderings of primitives. The color coding of primitives is consistent across different samples. 
}
%\vspace{-0.1cm}
\label{fig:ours_afhq}
\end{figure*}

\subsection{Training}
\label{subsec:training}
One way to train this framework is by learning the whole model end-to-end in an adversarial manner using only loss from an image-based discriminator similar to existing methods~\cite{piGAN2021}.
Our initial experiments in that direction proved to be unstable because of multiple moving components.
To stabilize training, we propose to train our model in a knowledge-distillation fashion using pre-trained EG3D~\cite{Chan2022} models.
The generators are trained in a supervised manner.
As EG3D can synthesize fairly consistent multi-view data for in-distribution camera views, we make use of it as the supervision.
Since the latent space of EG3D is known to be well structured, we mimic its properties by using their latent space $\rvw$ as $\rvw$ in our model.
Using EG3D intermediate latent space is critical in building a high-quality model as we will show later. 
Given multi-view renderings $\mathbf{X}$ and their corresponding latent space $\rvw$, we employ the following loss function to train our generators in a supervised manner.
\begin{align}
\mathcal{L} &= \mathcal{L_\text{rec}}+\mathcal{L_\text{disc}}+ \lambda_{perc}\mathcal{L_\text{perc}}
\end{align}
where $\mathcal{L_\text{rec}}$ is an $L_1$ reconstruction loss between rendered images from our model and $\mathbf{X}$. As reconstruction loss alone results in blurry results, we also employ a discriminator loss, which helps in getting sharper results. The discriminator loss is defined as follows,
\begin{align}
\mathcal{L_\text{disc}} &= f ( D(\mathcal{I}(\rvw) ) ) 
&+ f(-D(\mathbf{X})) + \lambda_{reg} \|\nabla D(\mathbf{X})\|^2 .
\label{eq:gan}
\end{align}
where $D$ is an image-based discriminator, $f(u) = -\log(1 + \exp(-u))$, and $\lambda_{reg}$ is the coefficient for $R_1$ regularization.

Although discriminator loss increases the sharpness of the results, it also introduces grid-like artifacts. We found having an additional perceptual loss helps in mitigating such effects. The perceptual loss $\mathcal{L_\text{perc}}$ is an LPIPS-based~\cite{zhang2018perceptual} loss between rendered images from our model and $\mathbf{X}$.

%%%%%%%%% BODY TEXT
\section{Results}

\begin{figure}

\newcommand{\lspace}[3]{ \hspace{.25cm} #1 \hspace{1.0cm}  #2  \hspace{1.0cm} #3  }

\centering
\includegraphics[width=0.4\textwidth]{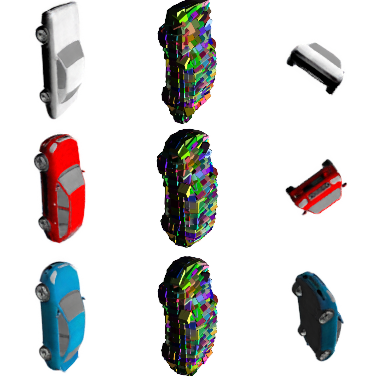}
%\vspace{-0.5cm}

\lspace{Sample}{Primitives}{Novel View}
\vspace{0.2cm}
\caption{Here we show samples of our method trained on the Shapenet Cars dataset. We also provide renderings of primitives. The color coding of primitives is consistent across different samples. Please note that our method is robust to different car shapes, even though we use a coarse fixed mesh as an initialization.  }
%\vspace{-0.1cm}
\label{fig:ours_cars}
\end{figure}

\paragraph{Datasets}
We demonstrate the results of our method on three datasets: FFHQ~\cite{Karras2019stylegan2}, AFHQv2-Cats~\cite{choi2020starganv2}, and Shapenet Cars~\cite{shapenet2015}. FFHQ~\cite{Karras2019stylegan2} is a large-scale portrait dataset of human faces under diverse camera poses. AFHQ-Cats~\cite{choi2020starganv2} is a similar dataset for cat portraits. Shapenet Cars~\cite{shapenet2015} is a synthetically rendered image dataset of different cars.

\paragraph{Baselines}
We compare our model with 3 baselines.
The first one is EG3D~\cite{Chan2022} as this is the state-of-the-art method that can synthesize high-resolution multi-view images in real-time.
We also train another model with the same architecture as our model, but with a learnable latent code for each sample. This is similar to auto-decoder training followed in LolNeRF~\cite{Rebain22} but trained with the dataset described above. We call this model MVP++. This experiment shows the importance of leveraging the $\rvw$ latent space in EG3D to train our models in a stable manner. 
Finally, we qualitatively compare to Authentic Volumetric Avatars~\cite{Cao22}. Although it~\cite{Cao22} is not a generative model, since they also use MVP~\cite{lombardi2021mixture} representation, we provide a qualitative comparison. Please note editing expressions is not a goal of our method, unlike~\cite{Cao22}.

\paragraph{Training Details}
We create a dataset of around 600k multi-view images and their latent spaces by sampling the pre-trained EG3D model on the FFHQ dataset. Similarly, we sample 300k images for the training model on AFHQ cats and Shapenet cars datasets.
Similar to MVP~\cite{lombardi2021mixture}, we also use opacity fade factor and volume minimization prior. The opacity fade factor encourages primitives to explain the scene content by the movement of primitives than to have primitives just modify their payload content. And the volume minimization prior makes the maximum usage of voxels to explain the opaque part of the scene rather than also including transparent parts. We refer readers to MVP~\cite{lombardi2021mixture} for more details.
We set $N_{prim}=1024$, $ \lambda_{perc}=20$ and $ \lambda_{reg}=0.0001$ and use mesh $\mathcal{M}$ with $1024$ vertices in all our experiments. The discriminator used for the discriminator loss is similar to the one used in $\pi$-GAN~\cite{piGAN2021} and is learned with a learning rate of 1e-5.
We train our models at $512\times512$ resolution with a batch size of 32 images for around 5 days on 4 Quadro RTX 8000 GPUs.
We use Adam optimizer~\cite{kingma2014adam} with a learning rate of $0.001$ for the parameters of the generator. 
Please refer to the supplemental document for more details.

\begin{figure*}
\newcommand{\lspace}[3]{\hspace{0.3cm} #1 \hspace{4.25cm}  #2  \hspace{4.25cm} #3  }
\centering
\includegraphics[width=1.0\textwidth]{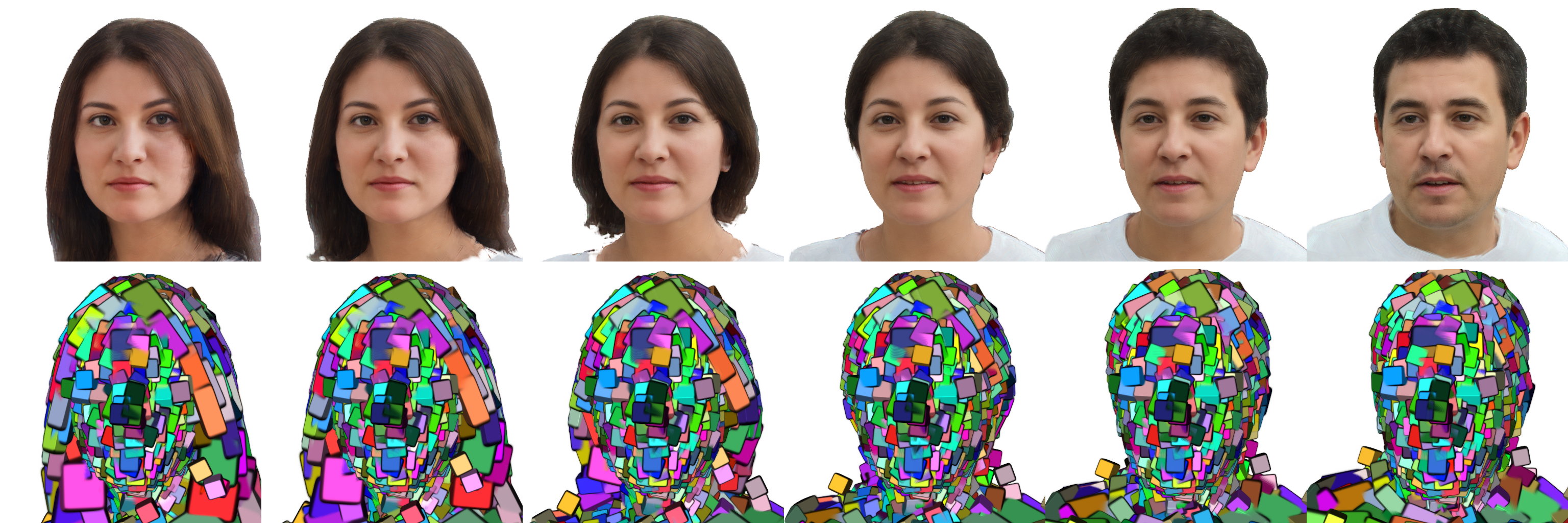}
%\vspace{-0.5cm}

\lspace{Sample1}{-----------Interpolation------------}{Sample2}

\caption{Here we show interpolation results of 2 random samples in the latent space. Even though our scene representation is represented by discrete primitives, we can observe how the primitives smoothly adapt themselves to synthesize high-quality results. For example, notice the primitive represented by an orange block covering a portion of long hair on the right-hand side adapts itself to change its position, orientation, and scale to finally represent a portion of short hair. This shows that our method captures correspondences across different identities. % can provide dense
}
%\vspace{-0.1cm}
\label{fig:ours_interpolation}
\end{figure*}

\begin{figure}

\newcommand{\lspace}[3]{ \hspace{-.5cm} #1 \hspace{1.0cm}  #2  \hspace{1.25cm} #3  }

\centering
\includegraphics[width=0.4\textwidth]{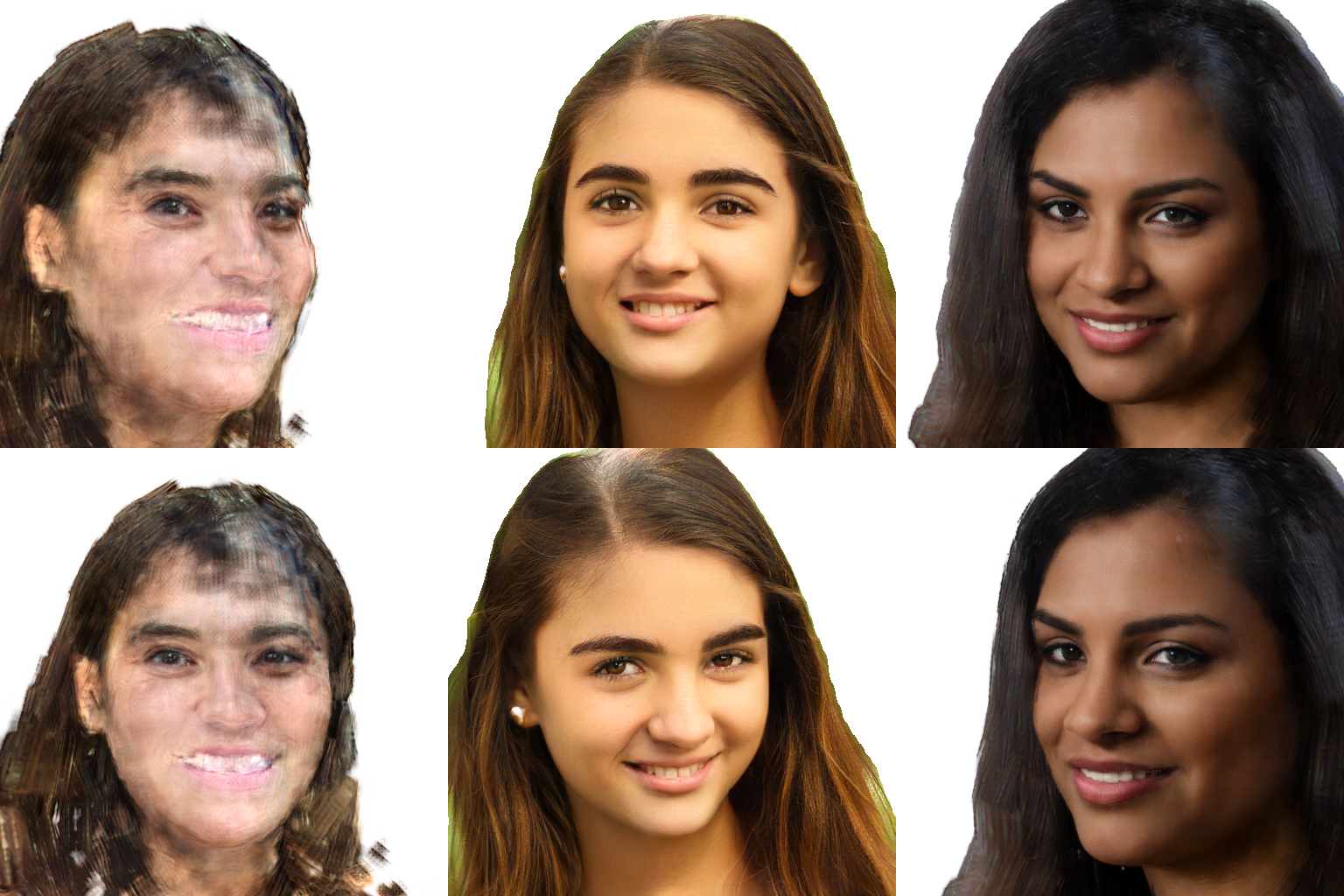}
%\vspace{-0.5cm}

\lspace{MVP++}{EG3D}{Ours}

\caption{Here, we compare our method to 2 baselines. We provide 2 novel view renderings of a random sample for all methods. MVP++ is trained in an auto-decoder manner, and as a result, has very poor random samples in its latent space. EG3D can synthesize high-quality  samples, but with the help of a 2D-based super-resolution module. In contrast, our method is synthesized with pure 3D representation and  is faster than EG3D.}
%\vspace{-0.1cm}
\label{fig:comp1}
\end{figure}

\begin{figure}

\newcommand{\lspace}[3]{ \hspace{1.0cm} #1 \hspace{0.85cm}  #2  \hspace{0.35cm} #3  }

\centering
\includegraphics[width=0.4\textwidth]{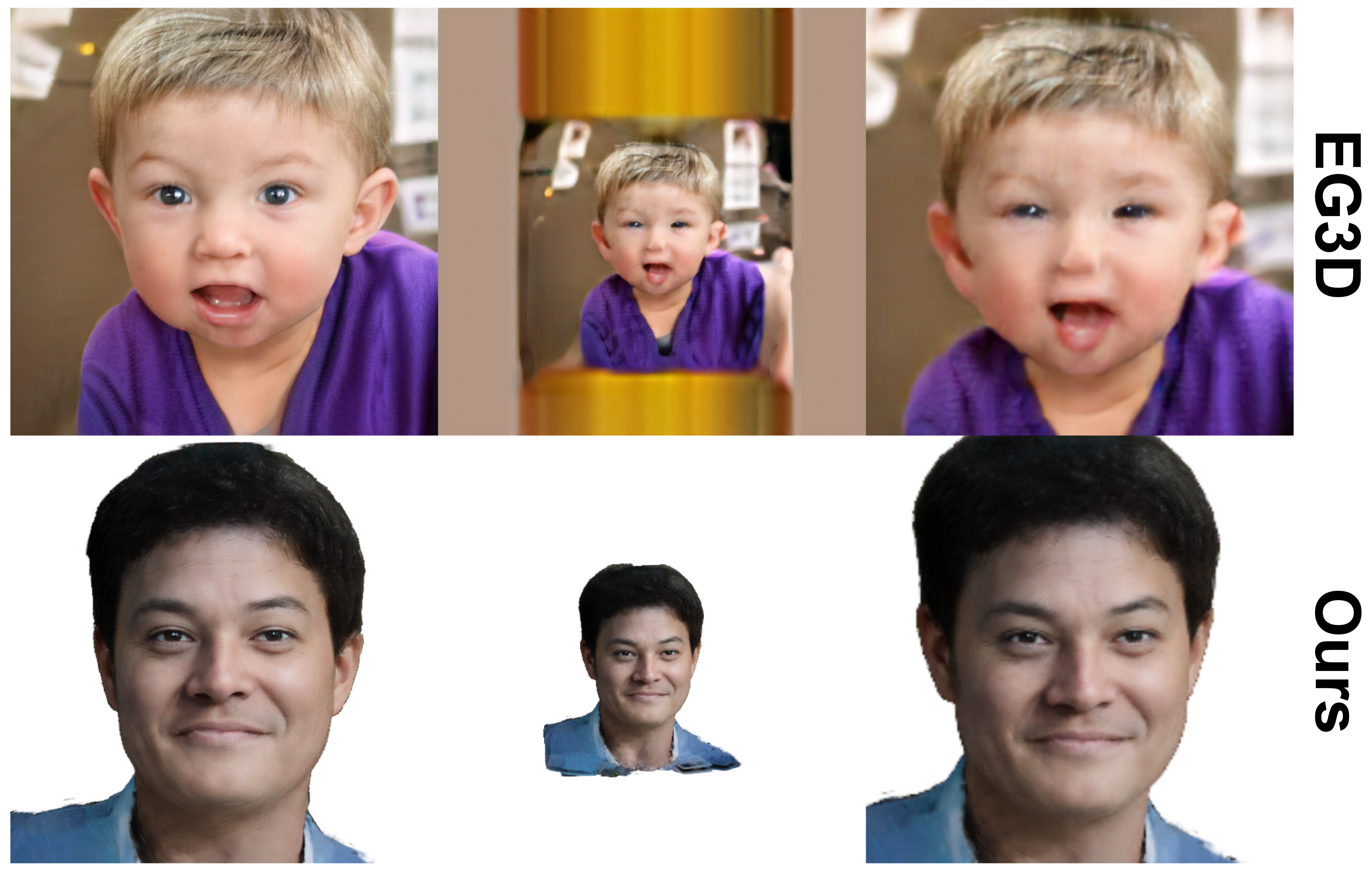}
%\vspace{-0.5cm}

\lspace{Sample}{Novel View}{Zoom(Novel View)}

\caption{ Here we compare our method (second row) with EG3D (first row) in out-of-training-distribution of camera pose. The first column shows samples for both our and EG3D in training distribution and the second column shows rendering with the camera placed a bit far and the third column shows the zoomed-in version of the second column. As can be seen, EG3D quality is not consistent. }
%\vspace{-0.1cm}
\label{fig:comp_far}
\end{figure}

\begin{figure}
\centering
\includegraphics[width=0.4\textwidth]{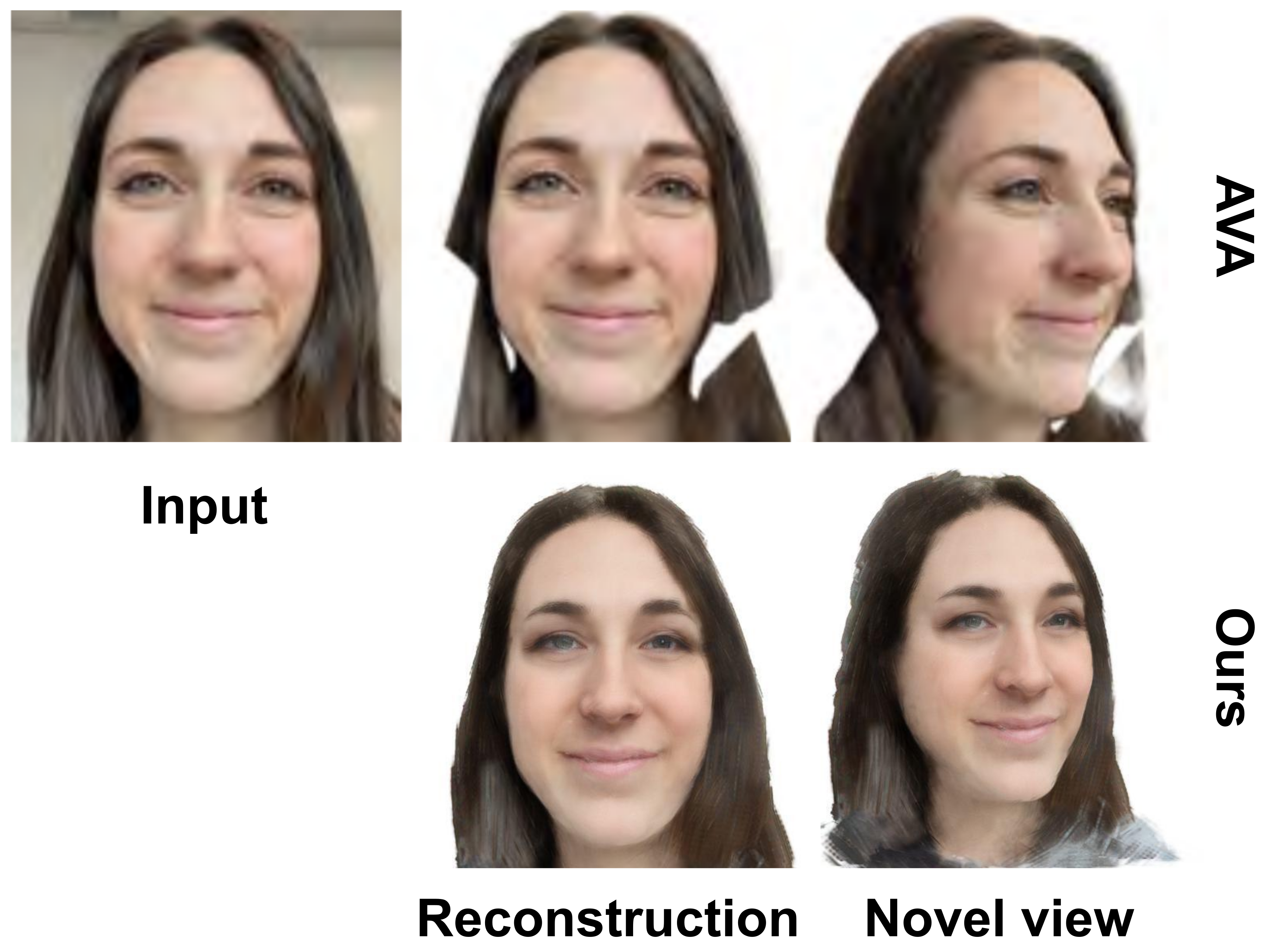}
%\vspace{-0.5cm}
\caption{ Here we show a comparison with Authentic Volumetric Avatar (AVA)~\cite{Cao22}. The first column shows the input image and the second column shows the reconstruction using AVA and ours. The third column shows the novel view rendering of both methods. As can be seen, AVA suffers from truncated hair reconstruction when reconstructing identity with long hair, while ours faithfully reconstructs long hair.}
%\vspace{-0.1cm}
\label{fig:comp_ava}
\end{figure}

\begin{table}[]
\centering
\begin{tabular}{lccc}
\toprule
     & MVP++ & EG3D & Ours \\
\midrule
%\cmidrule(lr){2-5}
FID (FFHQ)$\downarrow$ & 92.25     &    7.53          & 31.93       \\
Speed $\downarrow$ & 22.1     &    46.1           & 22.6       \\
\bottomrule
\end{tabular}
\caption{Quantitative comparisons using the FID score metric (a lower value is better) on the FFHQ dataset and computation time (in milliseconds). Here we see that ours perform better than MVP++ while EG3D performs the best in FID score. Our method is considerably faster than EG3D. Our method has other advantages of having pure 3D representation, rendering speed, and can provide correspondences across different samples. }
\label{tab:fid_ffhq}
%\vspace{-0.2cm}
\end{table}

\begin{table}[]
\centering
\begin{tabular}{lcc}
\toprule
     & EG3D & Ours \\
\midrule
%\cmidrule(lr){2-5}
FID (AFHQ) $\downarrow$    &    3.65          & 28.44       \\
%Speed     &    3.65          & 28.44       \\
\bottomrule
\end{tabular}
\caption{Quantitative comparisons using the FID score metric (a lower value is better) on the AFHQ dataset. EG3D performs better than our method on FID score, while our method has other advantages of having pure 3D representation, rendering speed, and can provide  correspondences across different samples. }
\label{tab:fid_afhq}
%\vspace{-0.2cm}
\end{table}

\begin{table}[]
\centering
\begin{tabular}{lcc}
\toprule
     & F1 (Out-Out) $\uparrow$ & F1 (Out-In) $\uparrow$ \\
\midrule
%\cmidrule(lr){2-5}
EG3D     &    0.2448          & 0.1832       \\
Ours     &    0.3333          & 0.2906       \\
\bottomrule
\end{tabular}
\caption{Quantitative comparisons using the F1 score of detected Facial Action Units (a higher value is better). Out-Out denotes the metric computed using renderings of the same sample under 2 views, which are both outside training-camera-pose-distribution. Out-In denotes the metric computed using renderings of the same sample under 2 views, with 1 view in training-camera-pose-distribution and the other being outside. Here we can see that our method clearly outperforms EG3D in both metrics.}
\label{tab:facs}
%\vspace{-0.2cm}
\end{table}

\paragraph{Qualitative Results}
We present the qualitative results of our method in Fig.~\ref{fig:ours_ffhq}, Fig.~\ref{fig:ours_afhq} and Fig.~\ref{fig:ours_cars} with models trained on FFHQ, AFHQ, and Shapenet car datasets respectively.
Our method can synthesize multi-view consistent results under different camera positions. It can also be observed that our method can generalize well to different categories of objects.
We also provide visualization of primitives rendering with consistent color coding across various samples. One can observe the correspondence between semantically similar parts of samples. 
In Fig.~\ref{fig:ours_interpolation}, we provide interpolation results between 2 randomly sampled faces. One can observe the smooth transition between them where the primitive move accordingly, which provides correspondence information that might be useful for downstream tasks. This is better appreciated in the video. Please refer to the supplemental video for the same.

Next, we provide a qualitative comparison of our method with 3 baselines. 
In Fig.~\ref{fig:comp1}, we compare our method with EG3D~\cite{Chan2022} and MVP++.
MPV++ has the same architecture and supervision as our method and  the only difference is that we use the  output of the mapping network of EG3D~\cite{Chan2022} as our latent space and MVP++ learns the latent vectors while training similar to LoLNeRF~\cite{Rebain22}. One can observe the random samples of MVP++ clearly suffer from severe artifacts. 
It is also observed numerically in Tab.~\ref{tab:fid_ffhq}.
The training strategy used in the Auto-Decoders framework results in sparse supervision of limited samples, equal to the number of training samples in latent space. In contrast, a model learned in an adversarial manner gets more supervision across dense points in latent space. As a result, the models trained in an adversarial manner are known to have well-behaved latent space.
Even though our model is not trained in a purely adversarial manner, as we use the latent space of EG3D, we clearly transfer good characteristics of the latent space of the adversarially trained model. 

We also provide a qualitative comparison with EG3D~\cite{Chan2022} in Fig.~\ref{fig:comp1}~\ref{fig:comp_far}. The samples from EG3D have slightly better perceptual quality than that of our method in capturing high-frequency details. This quality comes because of their 2D super-resolution module, which is known to capture high-frequency details better. In contrast, our representation is purely 3D without any 2D components.
This ensures multi-view consistency even for extreme out-of-distribution camera views.
In Fig.~\ref{fig:comp_far}, we provide renderings of both ours and EG3D in training camera distribution and a sample of rendering with a camera placed far from training distribution. As can be observed, EG3D suffers from degraded quality because of inconsistency in the 2D super-resolution module. %not a purely 3D representation. 
In contrast, our method retains the quality in out-of-training camera distribution.

In Fig.~\ref{fig:comp_ava}, we provide the monocular fitting results to a given input image. Given an input image, we use our trained model to fit to the input image and show reconstruction and novel view renderings. We obtain the camera pose for the input image using off-the-shelf face pose estimator~\cite{deng2019accurate}. Then we optimize the latent vector $w$ for about $1200$ iterations using reconstruction loss and simultaneously optimize the generator parameter and latent vector for another $800$ iterations to get the better fitting. In the same Fig.~\ref{fig:comp_ava}, we provide a comparison with Authentic Volumetric Avatars(AVA)~\cite{Cao22}. 
While AVA is not a generative model, it can generate a volumetric avatar using a phone scan, we thus qualitatively compare with this method as well.
Please note AVA is not a main comparison, as this method targets expression retargeting, while we don't. ~\cite{Cao22} heavily relies on the tracked mesh to place the primitives and suffers in the case of long hair, which results in truncated results. In contrast, our method is robust to long hair, illumination, and spectacles.  

\paragraph{Quantitative Results}
We provide commonly used FID scores to evaluate generative models in Tab.~\ref{tab:fid_ffhq}, ~\ref{tab:fid_afhq}. We sample 10k images for all the methods at $512 \times 512$ resolution. 
We can observe that EG3D~\cite{Chan2022} performs better than our model. Although perceptually samples from our model look good, we believe it suffers a higher score because of multiple reasons. Since our model has purely 3D representation without any 2D components, it is hard to achieve a very high-quality 2D synthesis for a generic model. Our model is also discontinuous in 3D space because of our efficient sparse primitive representation. This sometimes leads to piecewise linear regular patterns. 
But our model is robust to out-of-training-distribution camera renderings. To evaluate this scenario, we sample 1024 identities for both our model and EG3D and render them under 2 camera distributions. One distribution is similar to training data, as we can easily obtain that by using off-the-shelf face model estimators~\cite{deng2019accurate}. And the other is out of this distribution, by placing the camera a bit far from the face. 
To evaluate the multi-view consistency of these renderings, we obtain the active Facial Active Units using state-of-the-art detectors~\cite{luo2022learning} and compute the mean F1 score of all pairs of renderings for both methods. Our method clearly outperforms EG3D in this evaluation, proving the advantage of having pure 3D representation over a hybrid generator that relies on 2D super-resolution blocks. 

In spite of having pure 3D representation, our method can render images faster than EG3D. We report the average time taken to render $512 \times 512$ images for both methods in Tab.~\ref{tab:fid_ffhq}. Please note that the evaluation was done on an NVIDIA A40 GPU. The computation time for the EG3D model is calculated using their version which volumetrically renders at $64\times64$ and upsamples to $512\times512$. The version with $128\times128$ volumetric rendering is twice as slow as the one with $64\times64$.

%%%%%%%%% BODY TEXT
\section{Limitation}
Although our method can generate high-quality samples, it struggles in cases of curly hair. This is because of the discontinuous representation of the 3D scene. 
This is also shown numerically in the quantitative tables.
We also note that this representation couldn't be trained stably in a purely adversarial manner because of the discontinuous nature of the representation.

%%%%%%%%% BODY TEXT
\section{Conclusion}
We presented the first 3D-aware generative model with pure 3D scene representation that can be rendered at more than 43 FPS at $512\times512$ resolution.
In addition, our method can also provide dense correspondence between samples by tracking the primitive position, orientation, and scale.
Although our FID score is slightly worse than that of methods that rely on 2D super-resolution modules, our method has the advantage of robustness to camera views because of pure 3D representation, and computation speed and can provide correspondences.
We believe our method takes a solid step further in building high-quality, pure 3D, and efficient 3D-aware generative models and can inspire future work in this direction. 

\let\thefootnote\relax\footnotetext{
	\textbf{Acknowledgements:}
This work was supported by the ERC Consolidator Grant 4DReply (770784).
}

{\small
\bibliographystyle{ieee_fullname}
\bibliography{egbib}

\begin{thebibliography}{10}\itemsep=-1pt

\bibitem{bergman2022gnarf}
Alexander~W. Bergman, Petr Kellnhofer, Wang Yifan, Eric~R. Chan, David~B.
  Lindell, and Gordon Wetzstein.
\newblock Generative neural articulated radiance fields.
\newblock In {\em NeurIPS}, 2022.

\bibitem{Blanz99}
Volker Blanz and Thomas Vetter.
\newblock A morphable model for the synthesis of 3d faces.
\newblock In {\em Proceedings of the 26th Annual Conference on Computer
  Graphics and Interactive Techniques}, SIGGRAPH '99, page 187–194, USA,
  1999. ACM Press/Addison-Wesley Publishing Co.

\bibitem{Cao22}
Chen Cao, Tomas Simon, Jin~Kyu Kim, Gabe Schwartz, Michael Zollhoefer,
  Shun-Suke Saito, Stephen Lombardi, Shih-En Wei, Danielle Belko, Shoou-I Yu,
  Yaser Sheikh, and Jason Saragih.
\newblock Authentic volumetric avatars from a phone scan.
\newblock {\em ACM Trans. Graph.}, 41(4), jul 2022.

\bibitem{piGAN2021}
Eric Chan, Marco Monteiro, Petr Kellnhofer, Jiajun Wu, and Gordon Wetzstein.
\newblock pi-gan: Periodic implicit generative adversarial networks for
  3d-aware image synthesis.
\newblock In {\em Proc. CVPR}, 2021.

\bibitem{Chan2022}
Eric~R. Chan, Connor~Z. Lin, Matthew~A. Chan, Koki Nagano, Boxiao Pan,
  Shalini~De Mello, Orazio Gallo, Leonidas Guibas, Jonathan Tremblay, Sameh
  Khamis, Tero Karras, and Gordon Wetzstein.
\newblock Efficient geometry-aware {3D} generative adversarial networks.
\newblock In {\em CVPR}, 2022.

\bibitem{shapenet2015}
Angel~X. Chang, Thomas Funkhouser, Leonidas Guibas, Pat Hanrahan, Qixing Huang,
  Zimo Li, Silvio Savarese, Manolis Savva, Shuran Song, Hao Su, Jianxiong Xiao,
  Li Yi, and Fisher Yu.
\newblock {ShapeNet: An Information-Rich 3D Model Repository}.
\newblock Technical Report arXiv:1512.03012 [cs.GR], Stanford University ---
  Princeton University --- Toyota Technological Institute at Chicago, 2015.

\bibitem{choi2020starganv2}
Yunjey Choi, Youngjung Uh, Jaejun Yoo, and Jung-Woo Ha.
\newblock Stargan v2: Diverse image synthesis for multiple domains.
\newblock In {\em Proceedings of the IEEE Conference on Computer Vision and
  Pattern Recognition}, 2020.

\bibitem{blender}
Blender~Online Community.
\newblock {\em Blender - a 3D modelling and rendering package}.
\newblock Blender Foundation, Stichting Blender Foundation, Amsterdam, 2018.

\bibitem{corona2022lisa}
Enric Corona, Tomas Hodan, Minh Vo, Francesc Moreno-Noguer, Chris Sweeney,
  Richard Newcombe, and Lingni Ma.
\newblock Lisa: Learning implicit shape and appearance of hands.
\newblock In {\em CVPR}, 2022.

\bibitem{deng2022gram}
Yu Deng, Jiaolong Yang, Jianfeng Xiang, and Xin Tong.
\newblock Gram: Generative radiance manifolds for 3d-aware image generation.
\newblock In {\em IEEE/CVF Conference on Computer Vision and Pattern
  Recognition}, 2022.

\bibitem{deng2019accurate}
Yu Deng, Jiaolong Yang, Sicheng Xu, Dong Chen, Yunde Jia, and Xin Tong.
\newblock Accurate 3d face reconstruction with weakly-supervised learning: From
  single image to image set.
\newblock In {\em IEEE Computer Vision and Pattern Recognition Workshops},
  2019.

\bibitem{gu2022stylenerf}
Jiatao Gu, Lingjie Liu, Peng Wang, and Christian Theobalt.
\newblock Stylenerf: A style-based 3d aware generator for high-resolution image
  synthesis.
\newblock In {\em International Conference on Learning Representations}, 2022.

\bibitem{henzler2019escaping}
Philipp Henzler, Niloy~J Mitra, and Tobias Ritschel.
\newblock Escaping plato's cave: 3d shape from adversarial rendering.
\newblock In {\em Proceedings of the IEEE/CVF International Conference on
  Computer Vision}, pages 9984--9993, 2019.

\bibitem{Karras19}
Tero Karras, Samuli Laine, and Timo Aila.
\newblock A style-based generator architecture for generative adversarial
  networks.
\newblock In {\em 2019 IEEE/CVF Conference on Computer Vision and Pattern
  Recognition (CVPR)}, pages 4396--4405, 2019.

\bibitem{Karras2019stylegan2}
Tero Karras, Samuli Laine, Miika Aittala, Janne Hellsten, Jaakko Lehtinen, and
  Timo Aila.
\newblock Analyzing and improving the image quality of {StyleGAN}.
\newblock In {\em Proc. CVPR}, 2020.

\bibitem{kingma2014adam}
Diederik~P Kingma and Jimmy Ba.
\newblock Adam: A method for stochastic optimization.
\newblock {\em arXiv preprint arXiv:1412.6980}, 2014.

\bibitem{lombardi2021mixture}
Stephen Lombardi, Tomas Simon, Gabriel Schwartz, Michael Zollhoefer, Yaser
  Sheikh, and Jason Saragih.
\newblock Mixture of volumetric primitives for efficient neural rendering.
\newblock {\em ACM Transactions on Graphics (ToG)}, 40(4):1--13, 2021.

\bibitem{luo2022learning}
Cheng Luo, Siyang Song, Weicheng Xie, Linlin Shen, and Hatice Gunes.
\newblock Learning multi-dimensional edge feature-based au relation graph for
  facial action unit recognition.
\newblock In {\em Proceedings of the Thirty-First International Joint
  Conference on Artificial Intelligence, {IJCAI-22}}, pages 1239--1246, 2022.

\bibitem{mildenhall2020nerf}
Ben Mildenhall, Pratul~P. Srinivasan, Matthew Tancik, Jonathan~T. Barron, Ravi
  Ramamoorthi, and Ren Ng.
\newblock Nerf: Representing scenes as neural radiance fields for view
  synthesis.
\newblock In {\em ECCV}, 2020.

\bibitem{Niemeyer2020GIRAFFE}
Michael Niemeyer and Andreas Geiger.
\newblock Giraffe: Representing scenes as compositional generative neural
  feature fields.
\newblock In {\em Proc. IEEE Conf. on Computer Vision and Pattern Recognition
  (CVPR)}, 2021.

\bibitem{2021narf}
Atsuhiro Noguchi, Xiao Sun, Stephen Lin, and Tatsuya Harada.
\newblock Neural articulated radiance field.
\newblock In {\em International Conference on Computer Vision}, 2021.

\bibitem{orel2022styleSDF}
Roy Or-El, Xuan Luo, Mengyi Shan, Eli Shechtman, Jeong~Joon Park, and Ira
  Kemelmacher-Shlizerman.
\newblock Style{SDF}: {H}igh-{R}esolution {3D}-{C}onsistent {I}mage and
  {G}eometry {G}eneration.
\newblock In {\em Proceedings of the IEEE/CVF Conference on Computer Vision and
  Pattern Recognition (CVPR)}, pages 13503--13513, June 2022.

\bibitem{Park_2019_CVPR}
Jeong~Joon Park, Peter Florence, Julian Straub, Richard Newcombe, and Steven
  Lovegrove.
\newblock Deepsdf: Learning continuous signed distance functions for shape
  representation.
\newblock In {\em The IEEE Conference on Computer Vision and Pattern
  Recognition (CVPR)}, June 2019.

\bibitem{Paysan09}
Pascal Paysan, Reinhard Knothe, Brian Amberg, Sami Romdhani, and Thomas Vetter.
\newblock A 3d face model for pose and illumination invariant face recognition.
\newblock In {\em Proceedings of the 2009 Sixth IEEE International Conference
  on Advanced Video and Signal Based Surveillance}, AVSS '09, page 296–301,
  USA, 2009. IEEE Computer Society.

\bibitem{Rebain22}
Daniel Rebain, Mark~J. Matthews, Kwang Yi, Dmitry Lagun, and Andrea
  Tagliasacchi.
\newblock Lolnerf: Learn from one look.
\newblock In {\em Computer Vision Pattern Recognition (CVPR)}, 2022.

\bibitem{Schwarz2020NEURIPS}
Katja Schwarz, Yiyi Liao, Michael Niemeyer, and Andreas Geiger.
\newblock Graf: Generative radiance fields for 3d-aware image synthesis.
\newblock In {\em Advances in Neural Information Processing Systems (NeurIPS)},
  2020.

\bibitem{Schwarz2022NEURIPS}
Katja Schwarz, Axel Sauer, Michael Niemeyer, Yiyi Liao, and Andreas Geiger.
\newblock Voxgraf: Fast 3d-aware image synthesis with sparse voxel grids.
\newblock In {\em Advances in Neural Information Processing Systems (NeurIPS)},
  2022.

\bibitem{epigraf}
Ivan Skorokhodov, Sergey Tulyakov, Yiqun Wang, and Peter Wonka.
\newblock Epi{GRAF}: Rethinking training of 3d {GAN}s.
\newblock In Alice~H. Oh, Alekh Agarwal, Danielle Belgrave, and Kyunghyun Cho,
  editors, {\em Advances in Neural Information Processing Systems}, 2022.

\bibitem{Xiang22}
Jianfeng Xiang, Jiaolong Yang, Yu Deng, and Xin Tong.
\newblock Gram-hd: 3d-consistent image generation at high resolution with
  generative radiance manifolds, 2022.

\bibitem{xu2021volumegan}
Yinghao Xu, Sida Peng, Ceyuan Yang, Yujun Shen, and Bolei Zhou.
\newblock 3d-aware image synthesis via learning structural and textural
  representations.
\newblock {\em arXiv preprint arXiv:2112.10759}, 2021.

\bibitem{zhang2018perceptual}
Richard Zhang, Phillip Isola, Alexei~A Efros, Eli Shechtman, and Oliver Wang.
\newblock The unreasonable effectiveness of deep features as a perceptual
  metric.
\newblock In {\em CVPR}, 2018.

\bibitem{Zhang22}
Xuanmeng Zhang, Zhedong Zheng, Daiheng Gao, Bang Zhang, Pan Pan, and Yi Yang.
\newblock Multi-view consistent generative adversarial networks for 3d-aware
  image synthesis.
\newblock In {\em 2022 IEEE/CVF Conference on Computer Vision and Pattern
  Recognition (CVPR)}, pages 18429--18438, 2022.

\end{thebibliography}
}

\end{document}